# Extracting urban impervious surface from GF-1 imagery using one-class classifiers

**Authors:** Yao Yao*,[1], Jialv He[1], Jinbao Zhang[1], Yatao Zhang[2]
**Corresponding Author:** Yao Yao (yaoy33@mail2.sysu.edu.cn)
[1] School of Geography and Planning, Guangdong Key Laboratory for Urbanization and Geo-simulation, Sun Yat-sen University, Guangzhou, Guangdong province, China
[2] State key lab for information engineering in surveying mapping and remote sensing, Wuhan University, Wuhan, Hubei province, China

**Abstract:** Impervious surface area is a direct consequence of the urbanization, which also plays an important role in urban planning and environmental management. With the rapidly technical development of remote sensing, monitoring urban impervious surface via high spatial resolution (HSR) images has attracted unprecedented attention recently. Traditional multi-classes models are inefficient for impervious surface extraction because it requires labeling all needed and unneeded classes that occur in the image exhaustively. Therefore, we need to find a reliable one-class model to classify one specific land cover type without labeling other classes. In this study, we investigate several one-class classifiers, such as Presence and Background Learning (PBL), Positive Unlabeled Learning (PUL), OCSVM, BSVM and MAXENT, to extract urban impervious surface area using high spatial resolution imagery of GF-1, China's new generation of high spatial remote sensing satellite, and evaluate the classification accuracy based on artificial interpretation results. Compared to traditional multi-classes classifiers (ANN and SVM), the experimental results indicate that PBL and PUL provide higher classification accuracy, which is similar to the accuracy provided by ANN model. Meanwhile, PBL and PUL outperforms OCSVM, BSVM, MAXENT and SVM models. Hence, the one-class classifiers only need a small set of specific samples to train models without losing predictive accuracy, which is supposed to gain more attention on urban impervious surface extraction or other one specific land cover type.

# 1. Introduction

Impervious surface is a manufactured featured area where water cannot infiltrate into the soil, such as sidewalks, roads, driveways, parking lots, rooftops, and so on. In recent years, urban impervious surface has emerged as not only an indicator of the degree of urbanization, but also a major indicator of environmental quality (Arnold Jr and Gibbons 1996, Hong *et al.* 2014, Weng 2012). Therefore, it is of great significance to estimate and extract urban impervious surface areas in the researches of global climate change and human-environment interactions (Fan *et al.* 2013, Ma *et al.* 2014). In one word, the urban impervious surface area plays an important role in many urban issues, including environment management (Phinn *et al.* 2002, Weng *et al.* 2011), ecosystem analyses (Ridd 1995), urban planning (Lu and Weng 2006) and so on. Naturally, it is not strange that so many researchers have paid more concerns on impervious surface extraction (Deng *et al.* 2012, Deng and Wu 2013, Fan *et al.* 2013, Fan 2013, Fan and Deng 2014, Hong *et al.* 2013, Hong *et al.* 2014, Im *et al.* 2012, Zhang *et al.* 2014).

Remote sensing has been widely used in extracting and mapping urban impervious surface area. A number of methods used to classify mid-high resolution and high resolution images provide potential opportunities for detecting more-refined impervious surface information, including support vector machine (SVM), artificial neural network (ANN), multiple logistical regressions (MLR) and spectral mixture analysis (SMA) (Esch *et al.* 2009, Hu and Weng 2009, Im *et al.* 2012, Shao and Lunetta 2012, Sun *et al.* 2011, Wei and Blaschke 2014, Zhang *et al.* 2012, Zhang *et al.* 2014). However, these supervised classification methods need to label all land types in an image before training processes. If we use these multi-classification models to solve the one-class classification problem, such as impervious surface extraction, it will increase classification difficulty and time cost during manually labeling training samples, particularly in the case of using large mid-high or high resolution images (Li *et al.* 2011). But the fact tells us it is very necessary to extract impervious surface information from mid-high or high resolution remote sensing images.

A number of one-class models have been developed to solve one-class classification problem in the literature. In one-class classification models, the target class and other classes represent the class of interest and outliers respectively, and the models only need to label some target data (positive data) in training processes (Li 2013). Ratnaparkhi presented a method named Maximum Entropy Model (MAXENT or ME) (Ratnaparkhi and Others 1996), which has been successfully applied in estimating species geographic distributions (Phillips *et al.* 2006) and one-class classification of remote sensing imagery (Li and Guo 2010). The one-class support vector machine (OCSVM) method is another commonly used one-class classifier, which has shown promise in image classification and ecological niche modeling (Li *et al.* 2010, Song *et al.* 2008, Worner *et al.* 2014). The biased support vector machine (BSVM) (Liu *et al.* 2003), which is a state-of-the-art learning algorithm often used to solve unbalanced training data problems (Elkan and Noto 2008, Garg and Sundararajan 2009), has been studied in remote sensing one-class classification scenarios (Li *et al.* 2011). Besides, Li and Guo have developed one-class models named positive and unlabeled learning (PUL) and positive and background data learning (PBL) in succession. PUL has been proved to successfully obtain high classification accuracy of remote sensing images, and PBL has only been used to improve modeling of the geographical distributions of species(Li *et al.* 2011, Li *et al.* 2011). However, each of the proposed one-class classifiers has its shortcomings. MAXENT has a sophisticated mathematical model and it may cause a problem of parameters sensitivity and infinite iterations which lead to difficulty in realizing computer programming (Curran and Clark 2003). A drawback of SVM models, including the traditional SVM, OCSVM and BSVM, is sensitive to free parameters that are difficult to tune (Manevitz and Yousef 2002). PBL and PUL are successful in ecological niche modeling (Li *et al.* 2011, Li *et al.* 2011), which have rigorous mathematical models but few studies introduced the algorithm into high spatial remote sensing image classification.

In this study, we investigate several different one-class models (MAXENT, OCSVM, BSVM, PUL and PBL) and multi-classes models (ANN and SVM) to design several object-oriented classification experiments, aiming at extracting urban impervious surface area from image of GF-1, the China's new generation high spatial resolution satellite. Through assessing F-score, overall accuracy (OA), kappa coefficient ($\kappa$), user accuracy (UA) and product accuracy (PA) based on confusion matrixes and artificial interpolation, this paper presents horizontal comparisons of accuracies among classification results of these proposed classifiers. The results indicate that PBL and PUL classifiers provide the highest classification accuracy and reduce time cost of manually labeling data. They have nearly equal accuracy to ANN model, and outperform the OCSVM, BSVM, MAXENT and SVM models.

## 2. Methodology

## 2.1. Dataset Description

The dataset in our experiments is a mid-high spatial resolution (HSR) remote sensing image located in Guangzhou with 16 meters spatial resolution and size of 13669*13400 (1.36 GB).It was acquired in 2014 by wide field multispectral camera of GF-1 satellite, the China's new generation high resolution remote sensing satellite. Four spectral bands are available in GF-1 image: R(630-690nm), G(520-590nm), B(450-520nm), NIR(770-890nm). Using multi-scales segmentation method to segment image by ecognition 9.0, we extract 43,256 homogeneous image objects and 24 features, including spectral, shape and texture features, all of which were normalized to range [0, 1].

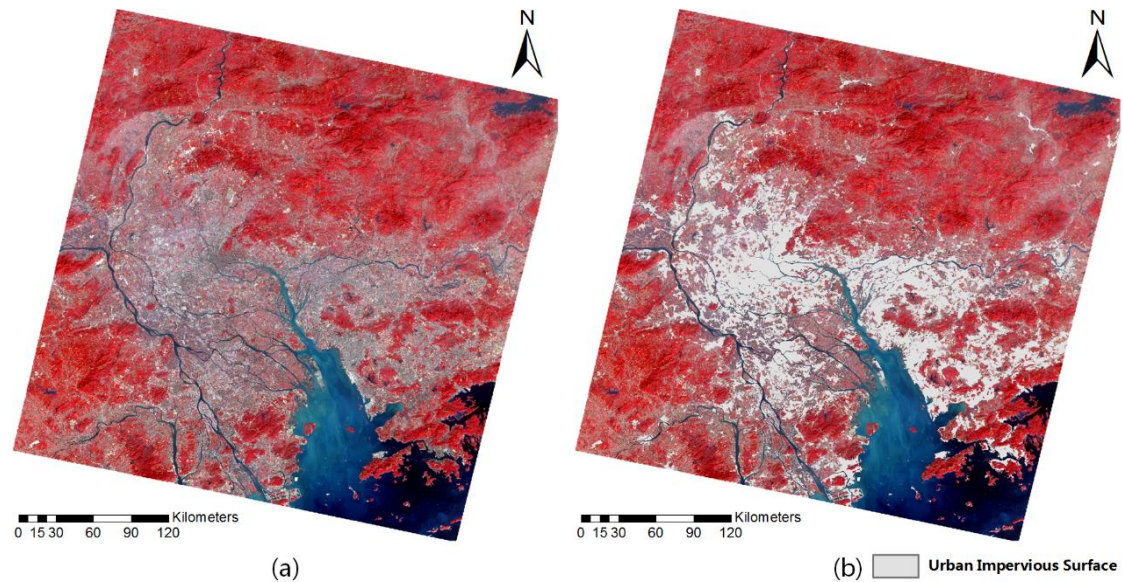

Figure 1: Original remote sensing image and interpretation result.

As shown in Figure 2, we first classified the image to urban impervious surface area and other areas by artificial interpretation. The interpretation result shows 7,574 objects were identified as urban impervious surface area. Based on artificial interpretation result, we randomly selected 1,000 objects of impervious surface area as positive training samples, and 15,000 objects, including impervious surface areas and others, as unlabeled (background) data for one-class classification, while another 15,000 objects of other areas are selected as negative samples for binary classification. The OCSVM only needs positive data for training, whereas PBL, PUL, MAXENT and BSVM require positive and background data. As comparing group, binary classification models (SVM and ANN) require positive and negative samples as input data in training processes.

In order to obtain statistically reliable accuracy assessment results, the training and predicting processes of each one-class and binary classifiers were tried for ten times using different groups of training dataset. And the classification results were evaluated by comparing with artificial interpretation results using F-score, overall accuracy (OA), kappa coefficient ($\kappa$), user accuracy (UA) and product accuracy (PA) .

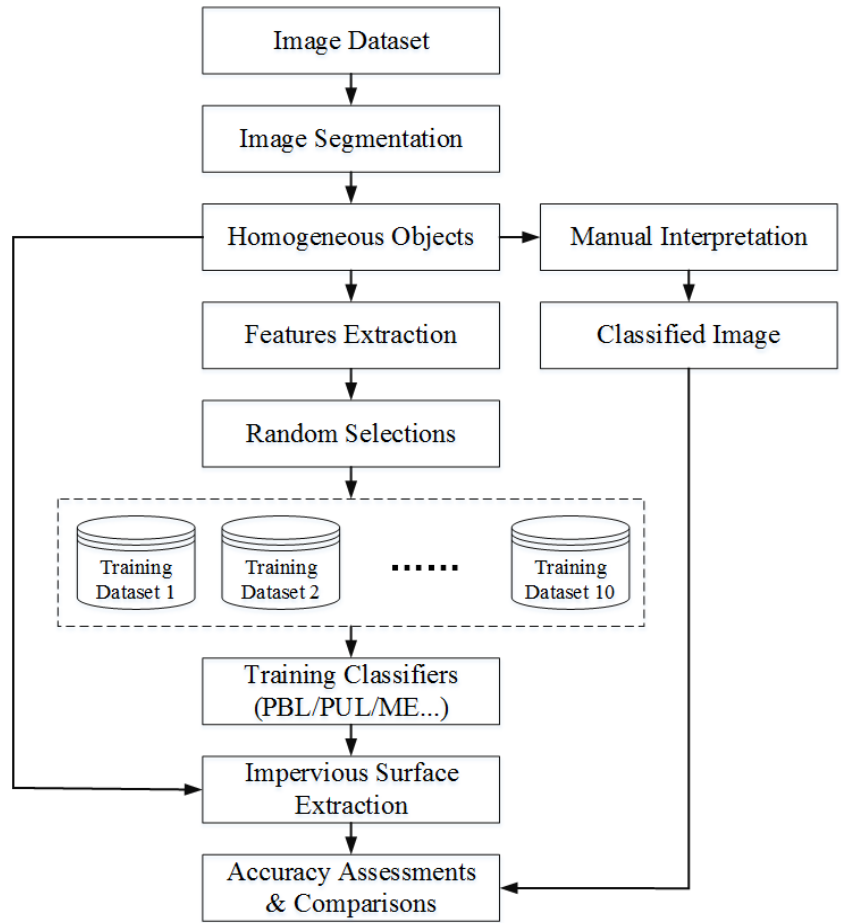

Figure 2: Work flow of extraction of urban impervious surface from GF-1 imagery based on one-class classifiers

In our experiments, we completed remote sensing segmentation and feature extraction by the ecognition 9.0, and implemented ANN, PBL, PUL, SVM models (including OCSVM, BSVM, SVM) by the Alglib 2.9 (http://www.alglib.net/) and libSVM 2.89 (http://www.csie.ntu.edu.tw/~cjlin/libsvm/) packages and C++ on VS2010 platform. One-class MAXENT results were calculated by a free maximum entropy modeling software for species geographic distributions of Princeton University (http://www.cs.princeton.edu/~schapire/maxent).

## 2.2. Model Developments

Positive-Unlabeled learning (PUL) and positive-background learning (PBL) were investigated by Li and Guo (Li *et al.* 2011, Li *et al.* 2011), which are proved effective in ecological niche modeling using positive and background dataset. We assumed that $x$ is denoted as environmental covariates, $y=1$ and $y=0$ are denoted as true presence and true absence, simultaneously $s=1$ and $s=0$ are denoted as observed presence and background data. And we defined that $p_1$ is the number of observed presence data. Background data contains $p_2$ presence data and $n_2$ absence data. $p_2$ and $n_2$ are unknown during the training process. Then we define $c = \frac{p_1}{p_1 + p_2}$ and $c$ can be estimated by:

$$c = \frac{1}{n}\sum_{x \in O} P(s=1|x,\eta=1) \qquad (1)$$

In the equation (1), $O$ means observed presences in training dataset. $n$ is the cardinality of $O$, and $\eta = 1$ means presence-background scenario. Then the existence probability $P(y=1|x)$ via PUL and PBL can be estimated by equation (2) and equation (3).

$$P(y=1|x) = \frac{P(s=1|x,\eta=1)}{c} \qquad (2)$$

$$P(y=1|x) = \frac{1-c}{c} \cdot \frac{P(s=1|x,\eta=1)}{1-P(s=1|x,\eta=1)} \qquad (3)$$

In equation (1) to (3), the conditional probability $P(s=1|x,\eta=1)$ can be estimated correctly by soft classifiers. Li have proved that PUL integrated with back-propagation neural network (BPNN) can estimate conditional probability and classifying image accurately (Li *et al.* 2011). In this paper, we use BPNN of 10 hidden layers implemented by Alglib package (Bochkanov and Bystritsky 2013) to train the PBL and PUL classifiers. Each BPNN classifier has been trained for 5 times and then the final prediction values were the average of all outputs so as to increase model reliability. After existence probability $P(y=1|x)$ was calculated, we set the existence probability threshold of impervious surface area as 0.5(explanation or not?).

As comparisons with methods above, one-class SVM (OCSVM) and biased SVM (BSVM) classifiers have been implemented by LIBSVM package (Chang and Lin 2011, Guo *et al.* 2008, Manevitz and Yousef 2002, Song *et al.* 2008). We segmented 25% of the training dataset as the validation dataset. Without negative data, F-score cannot be calculated, and $F_{pb}$ has been estimated as criteria of model parameters calibration. Two sensitive parameters of OCSVM with RGF kernel, penalty factor $C$ and kernel parameters $NU$, are need to be tuned, and $NU$ is an upper bound of the training error rate. We set $C \in [2^{-10}, 2^{10}]$ and $NU \in (0,1)$, and searched the optimum parameters by grid-search method, while the accuracy of OCSVM model is equaled to 1.0. For the BSVM classification approach, we need to tune three parameters: RBF kernel width $\gamma$, penalty factors of positive dataset $C+$ and penalty factor of negative dataset $C-$. The value of $\gamma$, $C+$ and $C-$ were searched the optimal values in the range of $[2^{-10}, 2^{10}]$ using grid-searching method. We trained the BSVM and tuned the model parameters using the whole training dataset including positive data and background data, while OCSVM only used the positive dataset.

In this paper, one class MAXENT software is freely provided by Princeton University (www.cs.princeton.edu/~schapire/maxent). In order to obtain the best estimation of presence probability, we used the logistical output format to train and predict the training data (positive and background data).. Free parameters and all homogeneous objects, which were the same as the data used in PBL classifiers, were set to the default values of software (Li *et al.* 2011, Phillips *et al.* 2006).

In order to estimate the effectiveness of these proposed one-class methods in accuracy assessment, we calculated the F-measure score ($F_{score}$), overall accuracy (OA), kappa coefficient ($\kappa$), user accuracy (UA) and product accuracy (PA) of each experiment based on confusion matrixes through comparing with interpretation image. $F_{pb}$ (proxy of F-measure based on positive–background data) could replace negative data with background data based on modified confusion matrix (Li and Guo 2014, Mack *et al.* 2014). And $F_{pb}$ is derived as follows (Li and Guo 2014):

$$F_{pb} = \frac{2 \times TP'}{TP' + FN' + FP'} \quad (4)$$

If we assumed that $s=1$ and $s=0$ means observed positive and background data respectively, moreover, $y'=1$ and $y'=0$ means classified positive and classified negative respectively, then in equation (4), $FP'$ means true-positive ($s=1$, $y'=1$), $FN'$ means false-positive ($s=0$, $y'=1$), and $TP'$ means false-negative ($s=1$, $y'=0$).

# 3. Results

In this study, through using 10 different datasets to train each classifier 10 times, we obtained the result accuracies of impervious surface area extraction compared with artificial interpretation result. What's more, we calculated the average values of accuracies of each classifier. Table 1 shows the confusion matrixes and average accuracy assessments results of all one-class classifiers, Figure 3 shows the box-plot images of F-score, $F_{pb}$, OA, kappa, commission error, omission error, PA and UA of all classifiers. The result images of impervious surface area extraction are shown in Figure 4.

Table 1 Confusion matrixes and average results of accuracy assessments of each one-class classifiers

| Methods | | Confusion Matrix | | Accuracy Assessments | | | | | |
|---|---|---|---|---|---|---|---|---|---|
| | | + | - | F-Score | F-pb | OA | Kappa | PA | UA |
| PBL | + | 0.8293 | 0.0328 | 0.7873 | 1.2676 | 0.9401 | 0.7524 | 0.7716 | 0.8036 |
| | - | 0.0271 | 0.1108 | | | | | 0.9684 | 0.962 |
| PUL | + | 0.8138 | 0.025 | 0.778 | 1.2759 | 0.9323 | 0.7382 | 0.8256 | 0.7355 |
| | - | 0.0426 | 0.1186 | | | | | 0.9502 | 0.9701 |
| MAXENT | + | 0.7919 | 0.0171 | 0.756 | 1.2099 | 0.9184 | 0.7082 | 0.8807 | 0.6623 |
| | - | 0.0645 | 0.1265 | | | | | 0.9247 | 0.9788 |
| OCSVM | + | 0.8259 | 0.0507 | 0.696 | 1.0662 | 0.9188 | 0.6494 | 0.6472 | 0.7527 |
| | - | 0.0305 | 0.0929 | | | | | 0.9644 | 0.9422 |
| BSVM | + | 0.5834 | 0.0002 | 0.5121 | 0.6803 | 0.7268 | 0.3796 | 0.9987 | 0.3443 |
| | - | 0.273 | 0.1434 | | | | | 0.6812 | 0.9997 |

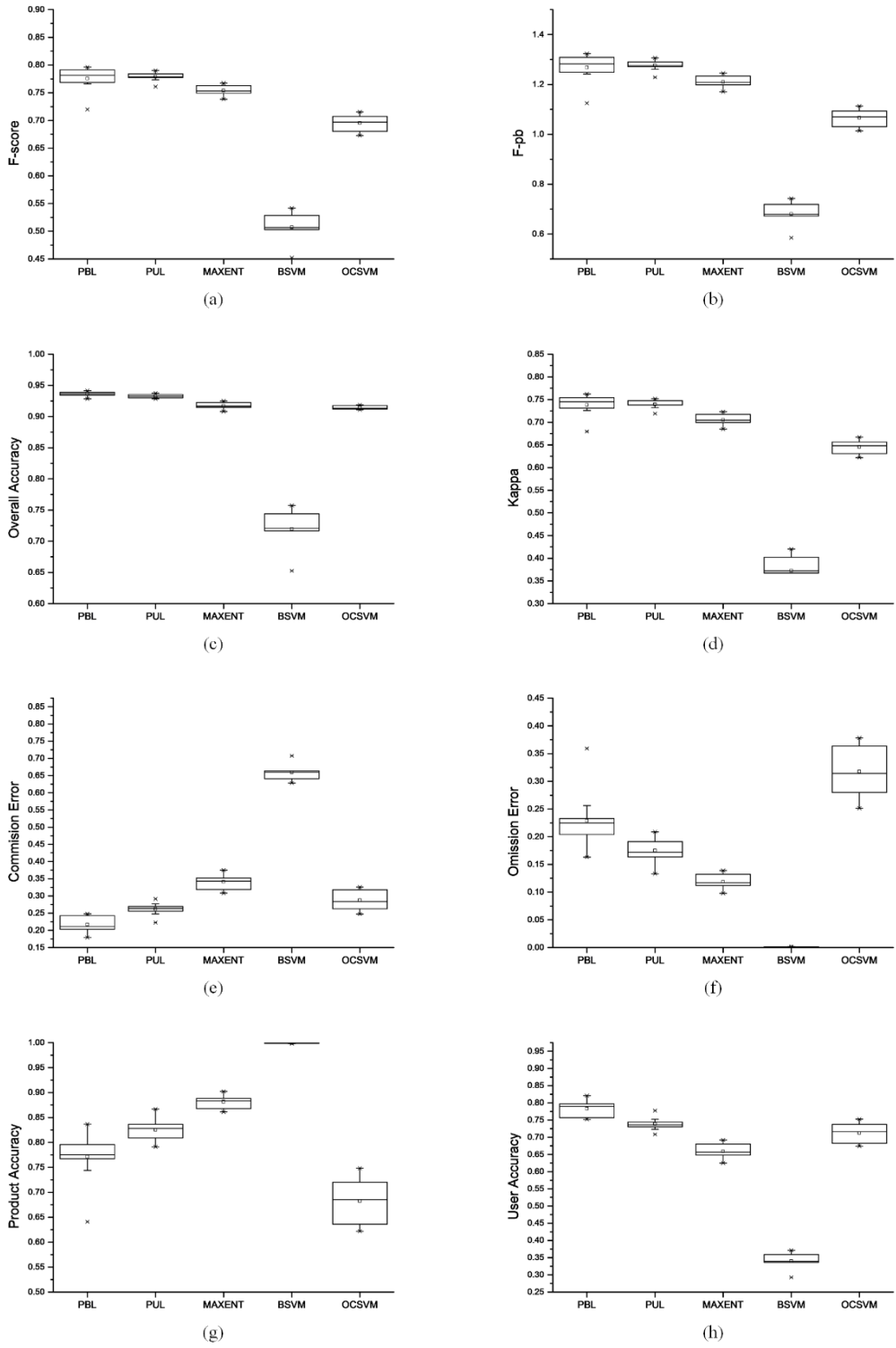

Figure 3: Accuracy assessment results of each one-class classifiers

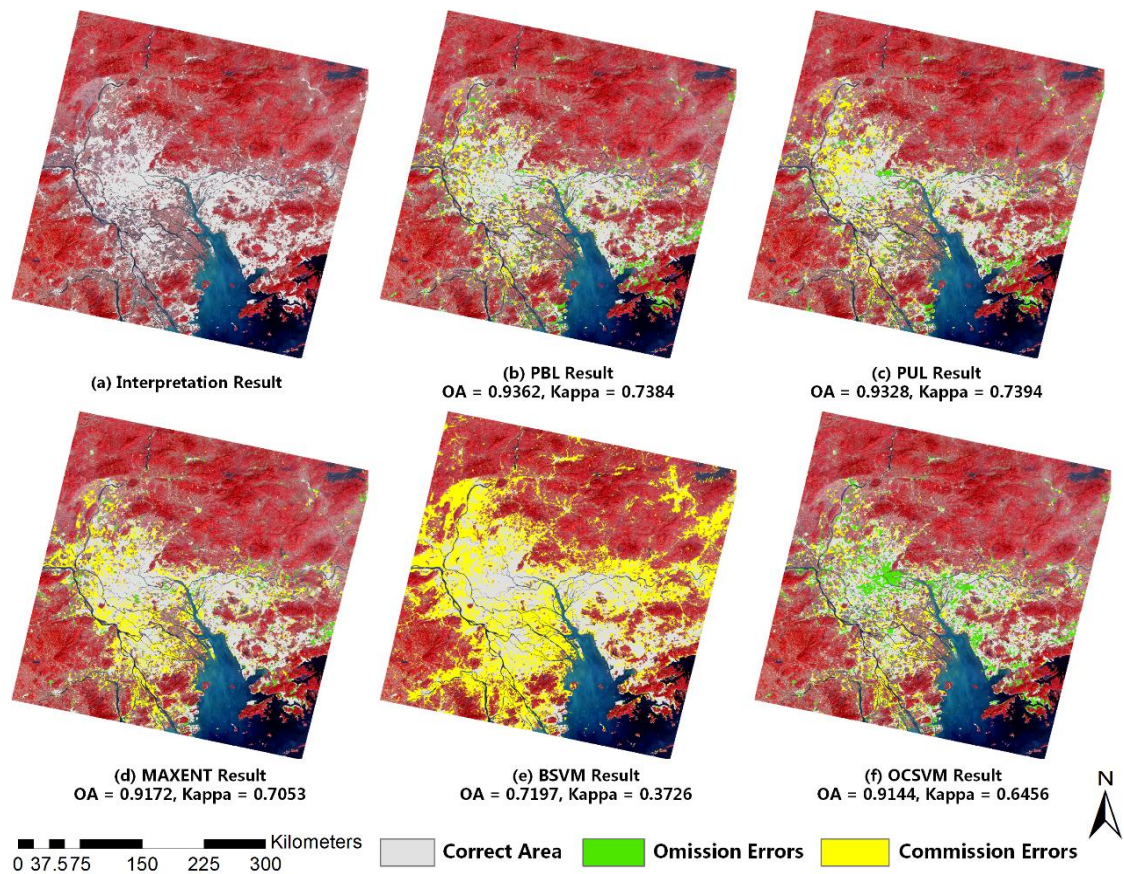

Figure 4: Impervious surface extraction results of each one-class classifiers. Compared with interpretation result, grey areas are correct, green and yellow areas are omission and commission errors.

Table 1 and Figure 3 display the average accuracies of each one-class classifier. As shown in Table 1, through comparing the impervious surface area extraction result of each classifier with artificial interpretation result, PBL and PUL obtained the highest F-score, $F_{pb}$, overall accuracy (OA) and kappa coefficient. And the classification performance of PBL is a little better than PUL. The classifiers using MAXENT and OCSVM also achieved good classification results, but the rather large omission error existed in the result of OCSVM, which caused lower accuracy estimation results of OCSVM. However, from figure 3 and figure 4, it is clear to see the results of BSVM obtain the worst classification accuracy, which should be due to severe commission error produced during the prediction period of BSVM. In general, impervious surface extraction using PBL classifier provides the highest average value of F-score, 0.7873, with a standard deviation of 0.0223. The performance of PUL is similar to PBL. The mean value and standard deviation of F-score are 0.7780 and 0.0080 respectively. Whereas the average F-score value of MAXENT, OCSVM and BSVM are 0.7560, 0.6960 and 0.5121, with standard deviations of 0.0100, 0.0279 and 0.0148.

# 4. Discussions

The area and distribution patterns of urban impervious surface are particularly key indicators to evaluate urbanization level and the urban environment. Urban impervious surface not only has a great effect on climate, ecology and environment of inner cities, but also has close connection with resident activities (Wouters 2014). Urban impervious surface area, which is developed by human activities in the performance of roads, driveways, sidewalks, roof, parking lots and so on, represents diversity, complexity and uncertainty on HSR remote sensing images. Thus, how to extract urban impervious surface area from high spatial resolution (HSR) remote sensing image accurately and efficiently is a huge challenge and an important problem to be solved.

In traditional multi-class classifiers, positive samples (urban impervious area) and negative samples (non-impervious surface area) have to be labeled simultaneously when urban impervious surface area is extracted. But it will cause a series of problems: on one hand, it is a vast workload and waste to label all categories on images, while the researchers are always interested in some special surface features, such as impervious surface area. On the other hand, due to the complexity of urban land surface, it is difficult to label non-impervious surface samples exactly, which will increase the manual selection errors in training dataset and reduce training and classification accuracies. Hence, in this research, we introduce one-class classification models which do not need negative samples during training processes to solve the problem of urban impervious surface area extraction from complicated urban land surface. These one-class classification models can obtain high classification accuracy as well as reduce the time cost of training dataset selection. During the training processes, OCSVM only needs positive samples, and the other one-class classifiers (PBL/PUL/MAXENT/BSVM) need amounts of unlabeled background samples. The real categories of background samples may belong to positive or negative samples, and it is simple to select these background samples completely at random in the whole of image.

In our research, we adopted the objects-oriented classification method to extract urban impervious surface area from GF-1 WFV image. Compared with the traditional pixel-based classification methods, the former method takes a full consideration of the structure and texture features of HSR image, and avoids the problem of random noise in pixel-level when calculating the average values of the homogeneous image objects during the feature extraction processes. It lowers the errors caused by remote sensing sensors especially in the area of accurate requirement matching (Zerrouki and

Bouchaffra 2014). In addition, because of the high spatial resolution of GF-1 WFV image, strong heterogeneity of urban impervious surface pixels and the random noises in HSR remote sensing image, the urban impervious surface extraction results of pixel-based classification methods may contain severe pepper and salt noises, while object-oriented classification methods can avoid producing these noises and have a better performance.

The PBL method is first introduced into the classification scenario of HSR sensing image. In this study, we employed the artificial neural network (ANN) classifier to estimate the existence probabilities $P(y=1|x)$ of urban impervious surface area on HSR image. Then we used a meaningful threshold (0.5 in this paper) cutting off the existence probabilities to obtain classification results. The results show that PBL-ANN method obtained the best classification accuracy of impervious surface area. Classification results of PUL approximate PBL due to the similar model framework, which trains a multi-classes classifier using presence-background dataset firstly, and then uses the $c$ value to correct the errors of the multi-classes classifier containing many training errors. As seen from classification results, in the case of presence-only situation, PBL is a more appropriate choice than PUL, especially for the one-class classification problem of HSR images.

Recent researches have indicated that MAXENT and OCSVM can obtain high classification accuracy even to 90% when used in one-class classification problems (Guo *et al.* 2008, Li and Guo 2010). The classification experiment results of MAXENT and OCSVM reflected that the overall accuracy of urban impervious surface area extraction could reach 91.84% and 91.88%, the kappa coefficient reached 0.7082 and 0.6494, and the corresponding F-score could reach 0.7560 and 0.6960 respectively. This means MAXENT and OCSVM can obtain high classification accuracy in one-class classification, but lower than PBL and PUL. However, during the process of implementation, the original mathematical model of MAXENT has amounts of iteration computation (Malouf 2002), so we used an approximate model with less iteration computation, which consumes less time on iterations yet to obtain higher accuracy (Figure 5).

In this paper, we set the Gaussian radical basis function (RBF) as the center kernel of SVM, OCSVM and BSVM, and tuned the sensitivity parameters (penalty factor $C$ and Kernel parameter $NU$) by particle swarm optimization (PSO), which also costs lots of time during the optimizing processes (Figure 5). And because SVM/OCSVM/BSVM classifiers have the problem of high sample sensitivity, the training time of different training dataset varies widely (Figure 5). In addition,

OCSVM only needs positive dataset to train classifier, and is greatly sensitive on outlier produced by manual samples selection, which leads to decreasing the accuracy and repeatability of classification results. Artificial neural network (ANN) were set as core classifier of PBL and PUL. As shown in Figure 5, we can see that the PBL and PUL had the shortest training time of all one-class classifiers and high fault tolerance to error samples, and the consuming time of different training dataset were consistent tolerably.

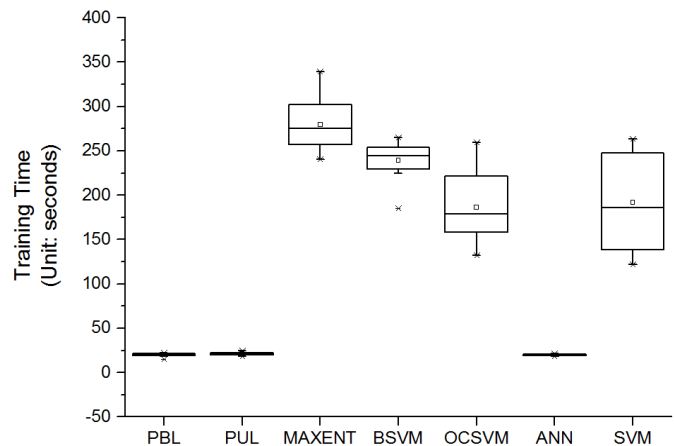

Figure 5: Comparisons of different classifiers using 10 different training datasets.

Interestingly, the classification accuracy of BSVM was the lowest. BSVM is a state-of-the-art classifier to solve one class classification problem using positive data and unlabeled data. However, it suffers several disadvantages. The first shortcoming of BSVM is the problem of high parameters sensitively. Therefore, we use the PSO algorithm to find the best parameters the same way as SVM and OCSVM, which causes problems of inefficient and instability during the training processes as shown in Figure 5. In addition, Figure 3 and Figure 4 show that BSVM suffered a serious problem of commission errors, and the reason is that the best usage scenario of BSVM is biased binary classification problem, meanwhile, BSVM classifier is sensitive to the missing data and error samples (Gonzalez-Abril *et al.* 2014). However, when BSVM was used in one-class classification scenario, the background datasets were selected randomly, which would contain an amount of non-negative samples. Training errors and commission errors would occur during the training and classification processes, which would lead to low accuracy of urban impervious surface area extraction using BSVM.

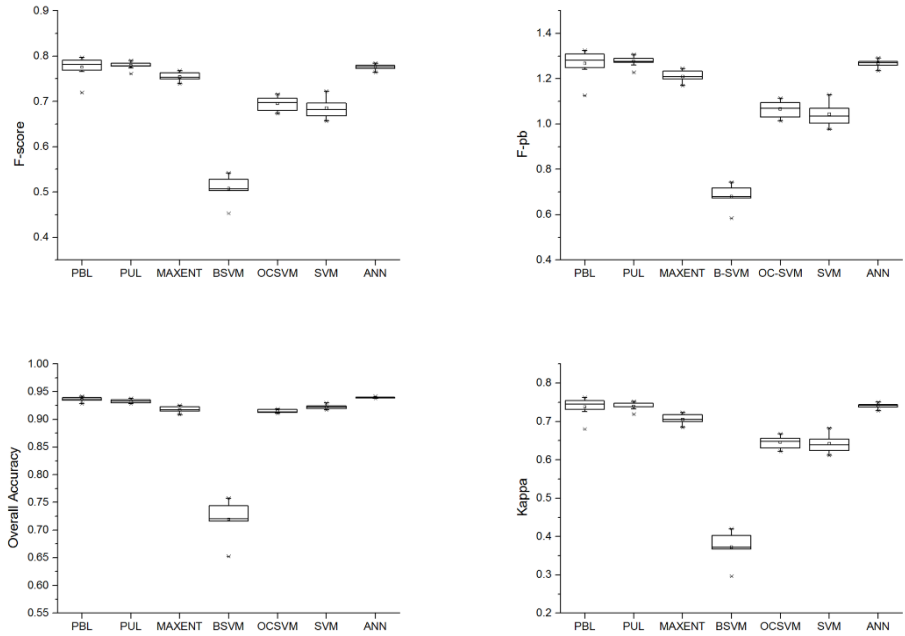

Figure 6: Accuracy comparisons of impervious surface area extraction with multi-class classifiers (ANN & SVM)

Furthermore, in this study, we also labeled 1,000 positive samples and 15,000 negative samples as the dataset to train traditional binary classifiers, such as SVM and ANN, and then evaluate the classification accuracy with the artificial interpretation. Through comparing with the classification results of one-class models, Figure 6 generally shows that, except for BSVM classifier, the overall classification accuracy of one-class models were slightly less than binary classifiers, but F-score, OA and kappa coefficient of PBL and PUL were similar with ANN, even higher than the classification results of SVM. In recent researches, Wanga and Ganga adopted spectral features and normalized difference impervious index (NDII) to extract impervious surface area from Landsat TM images of Nanjing, and OA and kappa reached 91.4% and 0.8 respectively (Patel and Mukherjee 2015). Hong and Yang extracted several linear ground features of Dianchi Lake from QuickBird using OOB method firstly, and then obtained urban impervious surface area by a fitting method named CART to fit the mixed land cover on Landsat images. As a result, the OA reached to 85.6% (Hong *et al.* 2014). Combining MODIS reflectance, MODIS NDVI and DMSP-OLS dataset, Lin and Liu adopted one-class MAXENT algorithm to extract urban land of China, the OA and kappa of which were 81% and 0.62 (Lin *et al.* 2014). So far, no research is effective to extract impervious surface areas through own spectral and texture features of HSR images exactly. However, the one-class classification methods, especially PBL and PUL, can obtain similar or higher accuracy of impervious surface areas extraction, even under the situation of positive samples existed only. The one-class model accustomed to one-class classification problem can effectively save time cost of manual samples selection, avoid the problems of parameters sensitivity and multiple iterations existing in MAXENT and SVM. Hence, it is possible to extract urban impervious surface areas from HSR images exactly and effectively by one-class models.

In addition, our proposed one-class models, including PBL and PUL, are inherited from the studies of Li and Guo (LiGuo and Elkan 2011c, LiGuo and Elkan 2011d). Meanwhile, this study has indicated that ANN-based PBL and PUL can obtain the highest training and classification accuracy. Moreover, PBL and PUL are flexible frameworks of one-class classification model, considering that any multi-class classifiers that can obtain the present probability can implement PBL and PUL. However, this study also has its own limitations that a large amount of samples is needed during the training processes of PBL and PUL, and we have barely probed into urban impervious surface areas extraction, though urban impervious surface area have the highest complexity and uncertainty on HSR remote sensing images. Future studies will determine the impact on PBL and PUL with more multi-class classifiers and the methods to decrease the sample size of training dataset.More case studies and evaluations are required.

## 5. Conclusions

In this paper, we adopted several one-class models, such as PBL, PUL, MAXENT, OCSVM and BSVM to extract urban impervious surface area of Guangzhou on GF-1 WFV image and compared with the results of artificial interpretation, multi-class models and previous researches. In general, one-class classification models integrating object-oriented classification method can extract urban impervious surface area from HSR remote sensing image successfully in the situation of positive samples existed only. Over all the one-class classifiers above, the ANN-based PBL and PUL can reach the optimum training efficiency and highest classification accuracy, meanwhile, the results of PBL and PUL are similar with or even superior to the results of multi-class classifiers. Hence, as a flexible one-class classification framework of HSR images, PBL and PUL reduce the time cost of manual label samples greatly but achieve high accuracy, which provides great research value on extracting special ground features from HSR images effectively and precisely. Therefore, we will construct the multi-scales ground features library of HSR images in the future work, and carry out research on more case studies of automatic extraction and precision validation of different special ground features in different areas.